\documentclass{article}

\usepackage{neurips_2023}

\usepackage[utf8]{inputenc} %
\usepackage[T1]{fontenc}    %
\usepackage{xcolor}
\definecolor{darkblue}{RGB}{0,0,112}
\usepackage[colorlinks=true,allcolors=darkblue]{hyperref}%
\usepackage{url}            %
\usepackage{booktabs}       %
\usepackage{amsfonts}       %
\usepackage{nicefrac}       %
\usepackage{microtype}      %
\usepackage{algorithm}
\usepackage{algpseudocode}
\usepackage{amsmath}
\usepackage{caption}
\usepackage{subcaption}
\usepackage{graphicx}
\usepackage{natbib}
\usepackage{svg}
\usepackage{wrapfig}
\usepackage{comment}
\setcitestyle{authoryear,open={},close={}}

\title{Reasoning with Latent Diffusion in Offline Reinforcement Learning}

\author{Siddarth Venkatraman\thanks{Equal contribution} \\
  Carnegie Mellon University, \\Mila, \& Universit\'e de Montr\'eal
  \And
  Shivesh Khaitan$^*$ \\
  Carnegie Mellon University
  \And
  Ravi Tej Akella$^*$ \\
  Carnegie Mellon University
  \And
  John Dolan \\
  Carnegie Mellon University
  \And
  Jeff Schneider \\
  Carnegie Mellon University
  \And
  Glen Berseth \\
  Mila, Universit\'e de Montr\'eal}

\begin{document}

\maketitle

\begin{abstract}

Offline reinforcement learning (RL) holds promise as a means to learn high-reward policies from a static dataset, without the need for further environment interactions. However, a key challenge in offline RL lies in effectively stitching portions of suboptimal trajectories from the static dataset while avoiding extrapolation errors arising due to a lack of support in the dataset. Existing approaches use conservative methods that are tricky to tune and struggle with multi-modal data (as we show) or rely on noisy Monte Carlo return-to-go samples for reward conditioning. In this work, we propose a novel approach that leverages the expressiveness of latent diffusion to model in-support trajectory sequences as compressed latent skills. This facilitates learning a Q-function while avoiding extrapolation error via batch-constraining. The latent space is also expressive and gracefully copes with multi-modal data. We show that the learned temporally-abstract latent space encodes richer task-specific information for offline RL tasks as compared to raw state-actions. This improves credit assignment and facilitates faster reward propagation during Q-learning. Our method demonstrates state-of-the-art performance on the D4RL benchmarks, particularly excelling in long-horizon, sparse-reward tasks.

\end{abstract}

\section{Introduction}

Offline reinforcement learning (RL) offers a promising approach to learning policies from static datasets. These datasets are often comprised of undirected demonstrations and suboptimal sequences collected using different \textit{behavior policies}.   
Several methods (\citet{fujimoto2019off, kostrikov2021offline, kumar2020conservative}) have been proposed for offline RL, all of which aim to strike a balance between constraining the learned policy to the support of the behavior policy and improving upon it. At the core of many of these approaches is an attempt to mitigate the \textit{extrapolation error} which arises while querying the learned Q-function on out-of-support samples for policy improvement. For example, in order to extract the best policy from the data, Q-learning uses an \textit{argmax}
over actions to obtain the temporal-difference target. However, querying the Q-function on out-of-support state-actions can lead to errors via exploiting an imperfect Q-function (\citet{fujimoto2019off}).

Framing offline RL as a generative modeling problem has gained significant traction (\citet{Chen2021DecisionTR, janner2021offline}); however, the performance is dependent on the power of the generative models used. These methods either avoid learning a Q-function or rely on other offline Q-learning methods. Recently diffusion models (\citet{pmlr-v37-sohl-dickstein15, song2019generative}), have emerged as state-of-the-art generative models for conditional image-generation (\citet{ramesh2022hierarchical, saharia2022photorealistic}). 
\textbf{Rather than avoiding Q-learning, we model the behavioral policy with diffusion and use this to avoid extrapolation error through batch-constraining.}
Previous diffusion-based sequence modeling methods in offline RL diffused over the raw state-action space. However, the low-level trajectory space tends to be poorly suited for reasoning. Some prior works (\citet{pmlr-v155-pertsch21a, ajay2020opal}) have proposed to instead reason in more well-conditioned spaces composed of higher-level behavioral primitives. Such temporal abstraction has been shown to result in faster and more reliable credit assignment (\citet{machado2023temporal, mann2014scaling}), particularly in long-horizon sparse-reward tasks.
\textbf{We harness the expressivity of powerful diffusion generative models to reason with temporal abstraction and improve credit assignment.}

Inspired by the recent successes of Latent Diffusion Models (LDMs) (\citet{rombach2022high, jun2023shap}), we propose learning similar latent trajectory representations for offline RL tasks by encoding rich high-level behaviors and learning a policy decoder to roll out low-level action sequences conditioned on these behaviors.
 The idea is to diffuse over semantically rich latent representations while relying on powerful decoders for high-frequency details. 
 Prior works which explored diffusion for offline RL (\citet{janner2022planning}, \citet{ajay2022conditional}) directly diffused over the raw state-action space, and their architectural considerations for effective diffusion models limited the networks to be simple U-Nets (\citet{inproceedings}). The separation of the diffusion model from the low-level policy allows us to model the low-level policy using a powerful autoregressive decoder. We perform state-conditioned latent diffusion on the learnt latent space and then learn a Q-function over states and corresponding latents. During Q-learning, we batch-constrain the candidate latents for the target Q-function using our expressive diffusion prior, thus mitigating extrapolation error. Our final policy samples latent skills from the LDM, scores the latents using the Q-function and executes the best behavior with the policy decoder. We refer to our method as \textbf{L}atent \textbf{D}iffusion-\textbf{C}onstrained \textbf{Q}-learning (LDCQ). Our method significantly improves results over previous offline RL methods, which suffer from extrapolation error or have difficulty in credit assignment in long-horizon sparse reward tasks.

\section{Related Work}

\textbf{Offline RL.} As discussed previously, offline RL poses the challenge of distributional shift while stitching suboptimal trajectories together. Conservative Q-Learning (CQL) (\citet{kumar2020conservative}) tries to constrain the policy to the behavioral support by learning a pessimistic Q-function that lower-bounds the optimal value function. Implicit Q-Learning (IQL) (\citet{pmlr-v151-vieillard22a}) tries to avoid extrapolation error by performing a trade-off between SARSA and DQN using expectile regression. However, it achieves the optimal batch-constrained policy only as their expectile parameter $\tau \to 1$, which leads to an increasingly difficult-to-optimize objective. Our method instead learns the optimal batch-constrained Q-function without introducing any pessimism or trade-off.  

Inspired by notable achievements of generative models in various domains including text-generation (\citet{vaswani2017attention}), speech synthesis (\cite{kong2020diffwave})
and image-generation (\citet{ramesh2022hierarchical, saharia2022photorealistic}), \citet{Chen2021DecisionTR} proposed to use a generative model for offline RL and bypass the need for Q-learning or bootstrapping altogether with \textit{return-conditioning} (\citet{srivastava2019training, kumar2019reward}). While these ideas have found success, getting a good return estimate for arbitrary states is not trivial and conditioning on returns outside the support of the training dataset can lead to the generative model producing low-value out-of-distribution sequences. Our method instead avoids return-conditioning and formulates a solution with batch-constraining which uses generative models to model the data distribution and use it to generate candidate actions to learn a Q-function without extrapolation-error (\citet{fujimoto2019off}). This formulation relies on the assumption that sampling from the generative model does not sample out-of-support samples, which has been difficult to achieve with previously used generative models in offline RL. Our method circumvents this problem with the latent diffusion model.
Further, to effectively address the problem of stitching, \citet{pmlr-v155-pertsch21a} and \citet{ajay2020opal} proposed learning policies in latent-trajectory spaces. However, they have to rely on a highly constrained latent space which is not rich enough for the downstream policy. This is due to the limitations of the generative model used like VAEs. Our proposed method to use latent diffusion, which can model complex distributions, allows for the needed flexibility in the latent space for effective Q-learning and the final policy. 

\textbf{Diffusion Probabilistic Models.} Recently, diffusion models (\citet{pmlr-v37-sohl-dickstein15, song2019generative}) have emerged as state-of-the-art generative models for conditional image-generation (\citet{ramesh2022hierarchical, saharia2022photorealistic}), super-resolution (\citet{saharia2022image}) and inpainting (\citet{lugmayr2022repaint}). They are a much more powerful class of generative model compared to Variational Autoencoders (VAEs) (\citet{kingma2013auto}), and benefit from a more stable training process as compared to Generative Adversarial Networks (GANs) (\cite{Goodfellow2014GenerativeAN}).
Recent works in offline RL (\citet{janner2022planning}, \citet{ajay2022conditional}) have proposed using diffusion to model trajectories and showcased its effectiveness in stitching together behaviors to improve upon suboptimal demonstrations. However, \citet{janner2022planning} make the assumption that the value function is learnt using other offline Q-learning methods and their classifier-guided diffusion requires querying the value function on noisy samples, which can lead to extrapolation-error. 
Similarly, \citet{ajay2022conditional} can suffer from distributional shift, as it relies on return-conditioning, and maximum returns from arbitrary states can be unknown without having access to a value function.
Our work proposes a method for learning Q-functions in latent trajectory space with latent diffusion while avoiding extrapolation-error and facilitating long horizon trajectory stitching and credit assignment.

\section{Preliminaries}
In this section, we introduce Diffusion Probabilistic Models and describe the Offline Reinforcement Learning problem on which we base our method.

\subsection{Diffusion Probabilistic Models}
Diffusion models (\citet{pmlr-v37-sohl-dickstein15, song2019generative}) are a class of latent variable generative model which learn to generate samples from a probability distribution $p(\bold{x})$ by mapping Gaussian noise to the target distribution through an iterative process. They are of the form  $p_\psi(\bold{x}_0) := \int p_\psi (\bold{x}_{0:T}) d\bold{x}_{1:T}$ where $\bold{x}_0, \dots \bold{x}_T$ are latent variables and the model defines the approximate posterior $q(\bold{x}_{1:T} \mid \bold{x}_0)$ through a fixed Markov chain which adds Gaussian noise to the data according to a variance schedule $\beta_1, \dots , \beta_T$. This process is called the \textit{forward diffusion process}:
\begin{equation}
    q(\bold{x}_{1:T} \mid \bold{x}_0) := \prod_{t=1}^{T} q(\bold{x}_t \mid \bold{x}_{t-1}), \qquad q(\bold{x}_t \mid \bold{x}_{t-1}) := \mathcal{N} (\bold{x}_t; \sqrt{1 - \beta_t} \bold{x}_{t-1}, \beta_t\bold{I})
\end{equation}
The forward distribution can be computed for an arbitrary timestep $t$ in closed form. Let $\alpha_t = 1 - \beta_t$ and $\bar{\alpha}_t = \prod_{i=1}^{t} \alpha_i$. Then $q(\bold{x}_t \mid \bold{x}_0) = \mathcal{N}(\bold{x}_t; \sqrt{\bar{\alpha}_t}\bold{x}_0, (1 - \bar{\alpha}_t) \bold{I})$.

Diffusion models learn to sample from the target distribution $p(\bold{x})$ by starting from Gaussian noise $p(\bold{x}_T) \sim \mathcal{N}(\bold{0}, \bold{I})$ and iteratively \textit{denoising} the noise to generate in-distribution samples. This is defined as the \textit{reverse diffusion process} $p_\psi(\bold{x}_{t-1} \mid \bold{x}_t)$:
\begin{equation}
p_\psi(\bold{x}_{0:T}) := p(\bold{x}_T)\prod_{t=1}^{T} p_\psi(\bold{x}_{t-1} \mid \bold{x}_t), \qquad p_\psi(\bold{x}_{t-1} \mid \bold{x}_t) := \mathcal{N} (\bold{x}_{t-1}; \bold{\mu}_\psi(\bold{x}_{t}, t), \bold{\Sigma}_\psi(\bold{x}_{t}, t))
\end{equation}
The reverse process is trained by minimizing a surrogate loss-function (\citet{ho2020denoising}):
\begin{equation}
    \mathcal{L}(\psi) = \mathbb{E}_{t \sim [1, T], \bold{x}_0 \sim q(\bold{x}_0), \epsilon \sim \mathcal{N}(\bold{0}, \bold{I})} \mid\mid \epsilon - \epsilon_\psi (\bold{x}_t, t) \mid\mid^2
\label{eqn:epsilonobj}
\end{equation}
Diffusion can be performed in a compressed latent space $\bold{z}$ (\citet{rombach2022high}) instead of the final high-dimensional output space of $\bold{x}$. This separates the reverse diffusion model $p_{\psi}(\bold{z}_{t-1} \mid \bold{z}_t)$ from the decoder $p_{\theta}(\bold{x} \mid \bold{z})$. The training is done in two stages, where the decoder is jointly trained with an encoder, similar to a $\beta$-Variational Autoencoder (\citet{kingma2013auto, Higgins2016betaVAELB}) with a low $\beta$. The prior is then trained to fit the optimized latents of this model. We explain this two-stage training in more detail in section \ref{two-stage}.

\subsection{Offline Reinforcement Learning}
The reinforcement learning (RL) problem can be formulated as a Markov decision process (MDP). This MDP is a tuple $\langle \rho_0, \mathcal{S}, \mathcal{A}, r, P, \gamma \rangle$, where $\rho_0$ is the initial state distribution,  $\mathcal{S}$ is a set of states, $\mathcal{A}$ is a set of actions, $r : \mathcal{S} \times \mathcal{A} \rightarrow \mathbb{R}$ is the reward function, $P : \mathcal{S} \times \mathcal{A} \times \mathcal{S} \rightarrow [0, 1]$ is the transition function that defines the probability of moving from one state to another after taking an action, and $\gamma \in [0, 1)$ is the discount factor that determines the importance of future rewards. The goal in RL is to learn a policy, i.e., a mapping from states to actions, that maximizes the expected cumulative discounted reward. In the offline RL setting (\citet{Levine2020OfflineRL}), the agent has access to a static dataset $\mathcal{D}=\{\bold{s}_t^i,\bold{a}_t^i,\bold{s}_{t+1}^i,r_t^i\}$ of transitions generated by a unknown behavior policy $\pi_{\beta}(\bold{a} \mid \bold{s})$ and the goal is to learn a new policy using only this dataset without interacting with the environment. Unlike behavioral cloning, offline RL methods seek to improve upon the behavior policy used to collect the offline dataset. The distribution mismatch between the behavior policy and the training policy can result in problems such as querying the target Q-function with actions not supported in the offline dataset leading to the extrapolation error problem. 

\section{Latent Diffusion Reinforcement Learning}
We begin by describing the two-stage training process for obtaining the low-level policy and the high-level latent diffusion prior. Next, we discuss how to use this prior to train a temporally abstract Q-function while avoiding bootstrapping error, and then use this Q-function during the policy execution phase. We finally describe an additional method to use the latent diffusion prior with goal-conditioning, which is more suitable for certain navigation tasks. All model architectures and hyperparameter choices are detailed in the supplemental material.

\subsection{Two-Stage LDM training}
\label{two-stage}
\textbf{Latent Representation and Low-Level Policy.} The first stage in training the latent diffusion model is comprised of learning a latent trajectory representation.
This means, given a dataset $\mathcal{D}$ of \textit{H}-length trajectories $\boldsymbol{\tau}_H$ represented as sequences of states and actions, $\bold{s}_0, \bold{a}_0, \bold{s}_1, \bold{a}_1, \cdots \bold{s}_{H - 1}, \bold{a}_{H - 1}$, we want to learn a low-level policy ${\pi_\theta}(\bold{a} \mspace{-4mu}\mid\mspace{-4mu} \bold{s}, \bold{z})$ such that $\bold{z}$ represents high-level behaviors in the trajectory. This is done using a $\beta$-Variational Autoencoder (VAE) (\citet{kingma2013auto, Higgins2016betaVAELB}). Specifically, we maximize the evidence lower bound (ELBO):
\begin{equation}
\mathcal{L}(\theta, \phi, \omega) = \mathbb{E}_{\boldsymbol{\tau}_H \sim D} [
\mathbb{E}_{q_\phi(\bold{z} \mid \boldsymbol{\tau}_H)}
[\sum_{t=0}^{H-1}\log \pi_\theta(\bold{a}_t \mspace{-2mu}\mid\mspace{-2mu} \bold{s}_t, \bold{z})] - \beta D_{KL}(q_\phi(\bold{z} \mspace{-2mu}\mid\mspace{-2mu} \boldsymbol{\tau}_H ) \mid\mid p_\omega(\bold{z} \mspace{-2mu}\mid\mspace{-2mu} \bold{s}_0))]
\end{equation}
where $q_\phi$ represents our approximate posterior over $\bold{z}$ given $\boldsymbol{\tau}_H$, and $p_\omega$ represents our conditional Gaussian prior over $\bold{z}$, given $\bold{s}_0$. Note that unlike BCQ, which uses a VAE's conditional Gaussian prior as the state-conditioned generative model, our latent diffusion model only uses the $\beta$-VAE to learn a latent space to diffuse over.
As such, the prior $p_{\omega}$ is simply a loose regularization of this latent space, and not a strong constraint. This is facilitated by the ability of latent diffusion models to later sample from such complex latent distributions. Prior works (\citet{pmlr-v155-pertsch21a, ajay2020opal}) have learned latent space representations of skills using VAEs. Their use of weaker Gaussian priors forces them to use higher values of the KL penalty multiplier $\beta$ to ensure the latents are well regularized. However, doing so restricts the information capacity of the latent, which limits the variation in behaviors captured by the latents. As we show in section \ref{temporal_abstraction}, increasing the horizon $H$ reveals a clear separation of useful behavioral modes when the latents are weakly constrained. 

The low-level policy $\pi_\theta$ is represented as an autoregressive model which can capture the fine details across the action dimensions, and is similar to the decoders used by \citet{ghasemipour2021emaq} and \citet{ajay2020opal}. While all the environments we test in this work use continuous action spaces, the use of latent diffusion allows the method to easily translate to discrete action spaces too, since the decoder can simply be altered to output a categorical distribution while the diffusion process remains unchanged.

\textbf{Latent Diffusion Prior.} For training the diffusion model, we collect a dataset of state-latent pairs ($\bold{s}_0$, $\bold{z}$) such that $\boldsymbol{\tau}_H \sim \mathcal{D}$ is a $H$-length trajectory snippet, $\bold{z} \sim q_{\phi}(\bold{z} \mspace{-2mu}\mid\mspace{-2mu} \boldsymbol{\tau}_H)$ where $q_{\phi}$ is the VAE encoder trained earlier, and $\bold{s}_0$ is the first state in $\boldsymbol{\tau}_H$. We want to model the prior $p(\bold{z} \mspace{-2mu}\mid\mspace{-2mu} \bold{s}_0)$, which is the distribution of the learnt latent space $\bold{z}$ conditioned on a state $\bold{s}_0$. This effectively represents the different behaviors possible from the state $\bold{s}_0$ as supported by the behavioral policy that collected the dataset. To this end, we learn a conditional latent diffusion model $p_\psi(\bold{z}\mspace{-2mu}\mid\mspace{-2mu} \bold{s}_0)$ by learning the time-dependent denoising function $\mu_{\psi}(\bold{z}_t, \bold{s}_0, t)$, which takes as input the current diffusion latent estimate $\bold{z}_t$ and the diffusion timestep $t$ to predict the original latent $\bold{z}_0$. Like \citet{ramesh2022hierarchical} and \citet{jun2023shap}, we found predicting the original latent $\bold{z}_0$ works better than predicting the noise $\boldsymbol{\epsilon}$. We reweigh the objective based on the noise level according to Min-SNR-$\gamma$ strategy (\citet{hang2023efficient}). This re-balances the objective, which otherwise is dominated by the loss terms corresponding to diffusion time steps closer to $T$. Concretely, we modify the objective in Eq. \ref{eqn:epsilonobj} to minimize:
\begin{equation}
    \mathcal{L}(\psi) = \mathbb{E}_{t \sim [1, T], \boldsymbol{\tau}_H \sim \mathcal{D},\bold{z}_0 \sim q_{\phi}(\bold{z} \mid \boldsymbol{\tau}_H), \bold{z}_t \sim q(\bold{z}_t \mid \bold{z}_0)} [\min\{\text{SNR}(t), \gamma\}( \mid\mid \bold{z}_0 - \mu_\psi (\bold{z}_t, \bold{s}_0,t) \mid\mid ^2)]
\end{equation}
Note that $q_{\phi}(\bold{z}\mid \boldsymbol{\tau}_H)$ is different from $q(\bold{z}_t\mid\bold{z}_0)$, the former being the approximate posterior of the trained VAE, while the latter is the forward Gaussian diffusion noising process. We use DDPM (\citet{ho2020denoising}) to sample from the diffusion prior in this work due to its simple implementation. As proposed in \citet{ho2022classifier}, we use classifier-free guidance for diffusion. This modifies the original training setup to learn both a conditional $\mu_\psi(\bold{z}_t, \bold{s}_0, t)$ and an unconditional model.
The unconditional version, is represented as $\mu_\psi(\bold{z}_t, $\O{}$ , t)$ where a dummy token \O{} takes the place of $\bold{s}_0$. The following update is then used to generate samples:
$\mu_\psi(\bold{z}_t, $\O{}$, t) + w (\mu_\psi(\bold{z}_t, \bold{s}_0, t) - \mu_\psi(\bold{z}_t, $\O{}$, t))$, where $w$ is a tunable hyper-parameter. Increasing $w$ results in reduced sample diversity, in favor of samples with high conditional density.

\begin{figure}[htpb]
    \centering
\includegraphics[width=\linewidth]{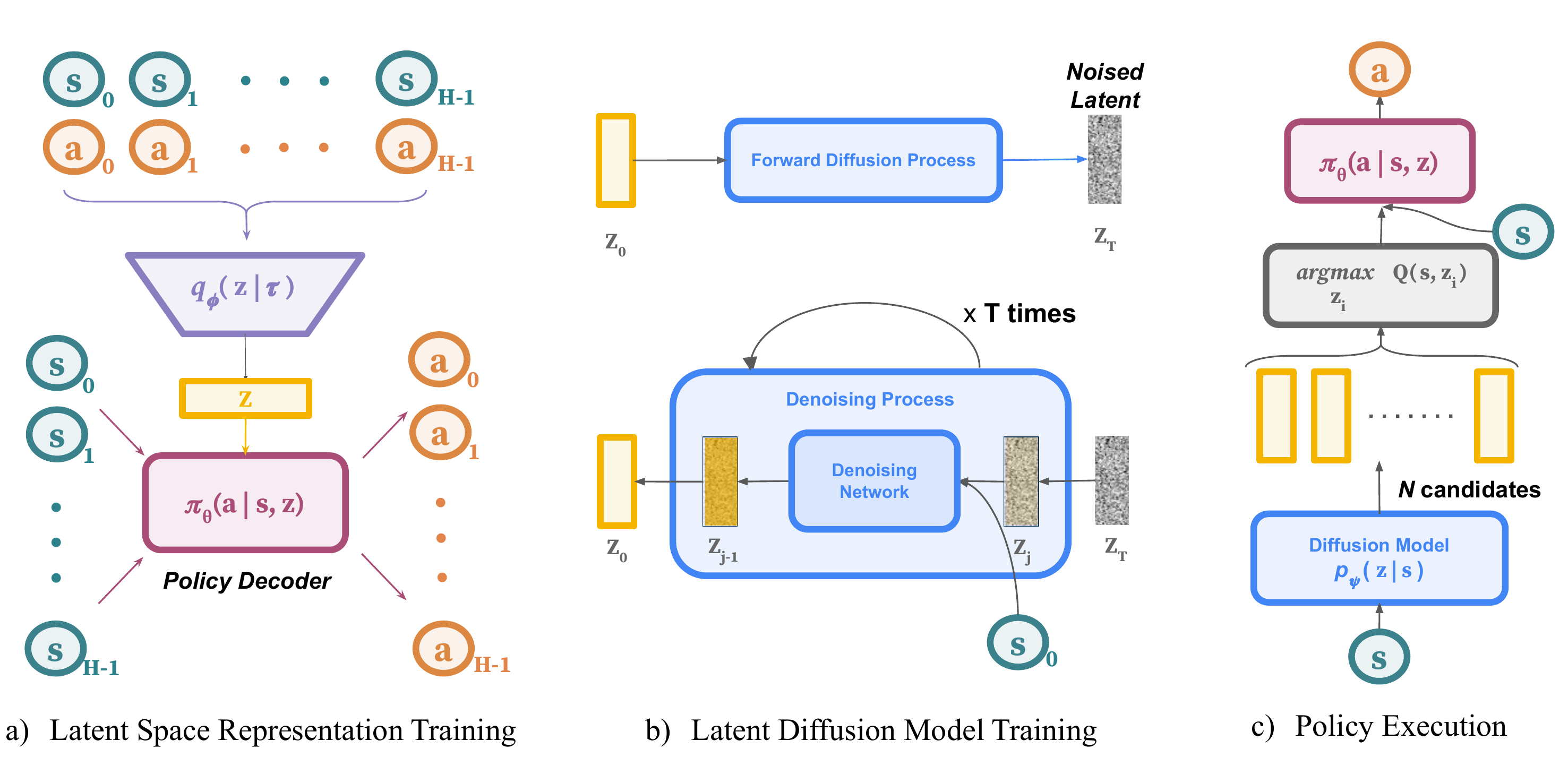}
    \caption{\textbf{Latent Diffusion Reinforcement Learning Overview} a) We first learn the latent space and low-level policy decoder by training a $\beta$-VAE over $H$-length sequences from the demonstrator dataset. b) We train a latent diffusion prior conditioned on $\bold{s}_0$ to predict latents generated by the VAE encoder. c) After learning a Q function using LDCQ (Algorithm \ref{alg:ldcq}), we score latents sampled by the prior with this Q function and execute the low-level policy $\pi_\theta$ conditioned on the argmax latent.}
    \label{fig:pipeline}
\end{figure}

\subsection{Latent Diffusion-Constrained Q-Learning (LDCQ)}

In batch-constrained Q-learning (BCQ), the target Q-function is constrained to only be maximized using actions that were taken by the demonstrator from the given state (\citet{fujimoto2019off}).
\begin{equation}
\pi(\bold{s}) = \underset{s.t.(\bold{s},\bold{a})\in \mathcal{D}}{\underset{\bold{a}}{\operatorname{argmax}}} Q(\bold{s},\bold{a})
\end{equation}
 In a deterministic MDP setting, BCQ is theoretically guaranteed to converge to the optimal batch-constrained policy. In any non-trivial setting, constraining the policy to actions having support from a given state in the dataset is not feasible, especially if the states are continuous. Instead, a function of the form $\pi_\psi(\bold{a} \mid \bold{s})$ must be learned on the demonstrator data and samples from this model are used as candidates for the argmax:
\begin{equation}
\pi(\bold{s}) = \underset{\bold{a}_i\sim \pi_\psi(\bold{a} \mid \bold{s})}{\operatorname{argmax}} Q(\bold{s},\bold{a}_i)
\end{equation}
However, in many offline RL datasets, the behavior policy is highly multimodal either due to the demonstrations being undirected, or because the behavior policy is actually a mixture of unimodal policies, making it difficult for previously used generative models like VAEs to sample from the distribution accurately.
The multimodality of this policy is further exacerbated with increases in temporal abstraction in the latent space, as we show in section \ref{temporal_abstraction}. We propose using latent diffusion to model this distribution, as diffusion is well suited for modelling such multi-modal distributions. We propose to learn a Q-function in the latent action space with latents sampled from the diffusion model. Specifically, we learn a Q-function $Q(\bold{s}, \bold{z})$, which represents the action-value of a latent action sequence $\bold{z}$ given state $\bold{s}$. At test time, we generate candidate latents from the diffusion prior $p_\psi(\bold{z} \mspace{-2mu}\mid\mspace{-2mu} \bold{s})$ and select the one which maximizes the learnt Q-function. We then use this latent with the low-level policy $\pi_\theta(\bold{a}_i \mspace{-2mu}\mid\mspace{-2mu} \bold{s}_i, \bold{z})$ to generate the action sequence for $H$ timesteps.

\textbf{Training.} We collect a replay buffer $\mathcal{B}$ for the dataset $\mathcal{D}$ of $H$-length trajectories and store transition tuples $(\bold{s}_t, \bold{z}, r_{t : t + H}, \bold{s}_{t + H})$ from $\boldsymbol{\tau}_H \sim \mathcal{D}$, where $\bold{s}_t$ is the first state in $\boldsymbol{\tau}_H$, $\bold{z} \sim q_{\phi}(\bold{z}  \mid \boldsymbol{\tau}_H)$ is the latent sampled from the VAE approximate posterior, $r_{t:t+H}$ represents the $\gamma$-discounted sum of rewards accumulated over the $H$ time-steps in $\boldsymbol{\tau}_H$, and $\bold{s}_{t + H}$ represents the state at the end of $H$-length trajectory snippet. The Q-function is learned with temporal-difference updates (\citet{sutton2018reinforcement}), where we sample a batch of latents for the target argmax using the diffusion prior $p_{\psi}(\bold{z} \mspace{-2mu}\mid\mspace{-2mu}  \bold{s}_{t + H})$. This should only sample latents which are under the support of the behavioral policy, and hence with the right temporal abstraction, allows for stitching skills to learn an optimal policy grounded on the data support. The resulting Q update can be summarized as:
\begin{equation}
Q(\bold{s}_t, \bold{z}) \leftarrow (r_{t : t + H} + \gamma^HQ(\bold{s}_{t + H}, \underset{\bold{z}_i \sim p_{\psi}(\bold{z} \mid \bold{s}_{t + H})}{\operatorname{argmax}}
 (Q(\bold{s}_{t + H}, \bold{z}_i))))
\end{equation}
We use Clipped Double Q-learning as proposed in (\citet{fujimoto2018addressing}) to further reduce overestimation bias during training. We also use Prioritized Experience Replay (\citet{schaul2015prioritized}) to accelerate the training in sparse-reward tasks like AntMaze and FrankaKitchen. We summarize our proposed LDCQ method in Algorithm \ref{alg:ldcq}.

\begin{algorithm}[htpb]
\caption{Latent Diffusion-Constrained Q-Learning (LDCQ)}
\label{alg:ldcq}
\begin{algorithmic}[1]
\State \textbf{Input:} prioritized-replay-buffer $\mathcal{B}$, horizon $H$, target network update-rate $\rho$, mini-batch size $N$, number of sampled latents $n$, maximum iterations $M$, discount-factor $\gamma$, latent diffusion denoising function $\mu_\psi$, variance schedule $\alpha_1, \dots , \alpha_T$, $\bar{\alpha}_1, \dots , \bar{\alpha}_T$, $\beta_1, \dots , \beta_T$.

\State Initialize Q-networks $Q_{\Theta_1}$ and $Q_{\Theta_2}$ with random parameters $Q_{\Theta_1}$, $Q_{\Theta_2}$ and target Q-networks $Q_{\Theta_{1}^{target}}$ and $Q_{\Theta_{2}^{target}}$ with $\Theta_{1}^{target} \leftarrow \Theta_{1}$, $\Theta_{2}^{target} \leftarrow \Theta_{2}$ 

\For{$iter=1$ to $M$}
    \State Sample a minibatch of $N$ transitions $\{(\bold{s}_t, \bold{z}, r_{t:t + H}, \bold{s}_{t + H})\}$ from $\mathcal{B}$

\State Sample $n$ latents for each transition: $\bold{z}_T \sim \mathcal{N(\bold{0}, \bold{I})}$

\For{$t=T$ to $1$} \Comment{DDPM Sampling}
\State $\hat{\bold{z}} = \mu_\psi(\bold{z}_t, $\O{}$, t) + w (\mu_\psi(\bold{z}_t, \bold{s}_{t + H}, t) - \mu_\psi(\bold{z}_t, $\O{}$, t))$

\State $\bold{z}_{t-1} \sim \mathcal{N}(\frac{\sqrt{\alpha_t}(1-\bar{\alpha}_{t-1})}{1 - \bar{\alpha}_t} \bold{z}_t + \frac{\sqrt{\bar{\alpha}_{t-1}}\beta_t}{1 - \bar{\alpha}_t} \hat{\bold{z}},
\mathbb{I} (t > 1)\beta_t\bold{I})$

\EndFor

    \State Compute the target values $y=r_{t:t+H}+ \gamma^H \{ \max\limits_{\bold{z}_0} \{ \min\limits_{j=1,2}  Q_{\Theta_{j}^{target}}(\bold{s}_{t + H}, \bold{z}_0) \} \}$
    \State Update $Q$-networks by minimizing the loss: $\frac{1}{N}||y -Q_\Theta(\bold{s}_t, \bold{z})||_2^2$
    \State Update target $Q$-networks: $\Theta^{target} \leftarrow \rho \Theta + (1 - \rho) \Theta^{target}$
\EndFor
\end{algorithmic}
\end{algorithm}

\textbf{Policy Execution.} The final policy for LDCQ comprises generating candidate latents $\bold{z}$ for a particular state $\bold{s}$ using the latent diffusion prior $\bold{z} \sim p_\psi(\bold{z} \mid \bold{s})$. These latents are then scored using the learnt Q-function and the best latent $\bold{z}_{max}$ is decoded using the VAE autoregressive decoder $\bold{a} \sim \pi_\theta(\bold{a} \mid \bold{s}, \bold{z}_{max})$ to obtain H-length action sequences which are executed sequentially. Note that the latent diffusion model is used both during training the Q-function and during the final evaluation phase, ensuring that the sampled latents do not go out-of-support.

\subsection{Latent Diffusion Goal Conditioning (LDGC)}
Diffuser (\citet{janner2022planning}) proposed framing certain navigation problems as a sequence inpainting task, where the last state of the diffused trajectory is set to be the goal during sampling. We can similarly condition our diffusion prior on the goal to sample from feasible latents that lead to the goal. This prior is of the form $p_{\psi}(\bold{z} \mspace{-2mu}\mid\mspace{-2mu}  \bold{s}_0, \bold{s}_g)$, where $\bold{s}_g$ is the target goal state. Since with latent diffusion, the training of the low-level policy alongside the VAE is done separately from the diffusion prior training, we can reuse the same VAE posterior to train different diffusion models, such as this goal-conditioned variant. At test time, we perform classifer-free guidance to further push the sampling towards high-density goal-conditioned latents. For tasks which are suited to goal conditioning, this can be simpler to implement and achieves better performance than Q-learning. Also, unlike Diffuser, our method does not need to have the goal within the planning horizon of the trajectory. This allows our method to be used for arbitrarily long-horizon tasks.

\section{Experimental Evaluation and Analysis}

In our experiments, we focus on \textbf{1)} studying the effect of temporal abstraction on the latent space (section \ref{temporal_abstraction}) \textbf{2)} understanding the need for diffusion to model the latent space (section \ref{sec:LDMmultimodal} and \ref{sec:performanceimprovement}) and \textbf{3)} evaluating the performance of our method in the D4RL offline RL benchmarks (section \ref{sec:offlinerl}).

\subsection{Temporal abstraction induces multi-modality in latent space}
\label{temporal_abstraction}
In this section, we study how the horizon length $H$ affects the latent space and provide empirical justification for learning long-horizon latent space representations. For our experiment, we consider the \textit{kitchen-mixed-v0} task from the D4RL benchmark suite (\citet{fu2020d4rl}).
The goal in this task is to control a 9-DoF robotic arm to manipulate multiple objects (microwave, kettle, burner and a switch) sequentially, in a single episode to reach a desired configuration, with only sparse 0-1 completion reward for every object that attains the target configuration. As \citet{fu2020d4rl} states, there is a high degree of multi-modality in this task arising from the demonstration trajectories because different trajectories in the dataset complete the tasks in a random order. Thus, before starting to solve any task, the policy implicitly needs to \textit{choose} which task to solve and then generate the actions to solve the task. Given a state, the dataset can consist of multiple behavior modes, and averaging over these modes leads to suboptimal action sequences. Hence, being able to differentiate between these tasks is desirable.

\begin{figure}[ht]
    \centering
        \includegraphics[width=\linewidth]{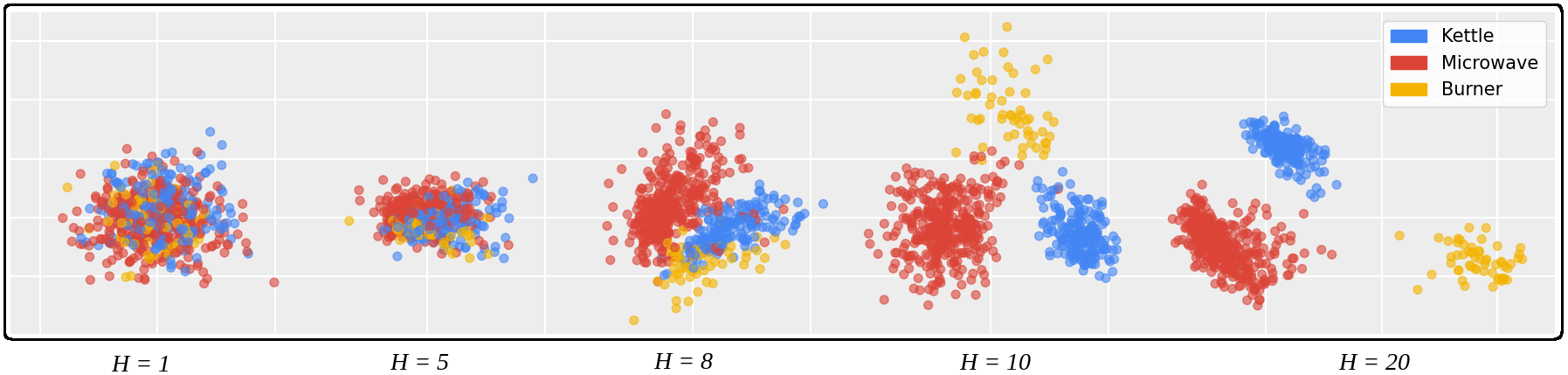}
    \caption{\textbf{Projection of latents across horizon}. Latent projections of trajectory snippets with different horizon lengths $H$. From the initial state there are 3 tasks (Kettle, Microwave, Burner) which are randomly selected at the start of each episode. These 3 primary modes emerge as we increase $H$, with the distribution turning multi-modal.}
    \label{fig:latents_horizons}
\end{figure}

We hypothesize that as we increase our sequence horizon $H$, we should see better separation between the modes. In \textcolor{darkblue}{Figure \ref{fig:latents_horizons}},
we plot a 2D (PCA) projection of the VAE encoder latents of the starting state-action sequences in the kitchen-mixed dataset. With a lower horizon, these modes are difficult to isolate and the latents appear to be drawn from a Normal distribution (\textcolor{darkblue}{Figure \ref{fig:latents_horizons}}). However, as we increase temporal abstraction from $H=1$ to $H=20$, we can see \textit{three} distinct modes emerge, which when cross-referenced with the dataset correspond to the three common tasks executed from the starting state by the behavioral policy (microwave, kettle, and burner). 
These modes capture underlying variation in an action sequence, and having picked one we can run our low-level policy to execute it. As demonstrated in our experiments, such temporal abstraction facilitates easier Q-stitching, with better asymptotic performance. However, in order to train these abstract Q functions, it becomes necessary to sample from the complex multi-modal distribution and the conventional VAE conditional Gaussian prior is no longer adequate for this purpose, as shown in section \ref{sec:LDMmultimodal}.

\subsection{LDMs address multi-modality in latent space}
\label{sec:LDMmultimodal}
\begin{figure}[ht]
    \centering
    \begin{subfigure}[b]{0.3\linewidth}
        \centering
        \includegraphics[width=\linewidth]{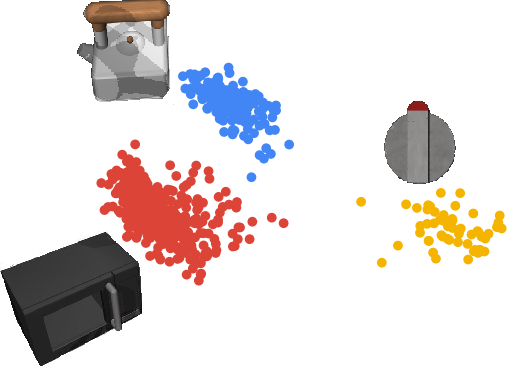}
        \caption{Ground truth}
        \label{fig:kitchenlatentsvizgt}
    \end{subfigure}
    \quad
    \begin{subfigure}[b]{0.3\linewidth}
        \centering
        \includegraphics[width=\linewidth]{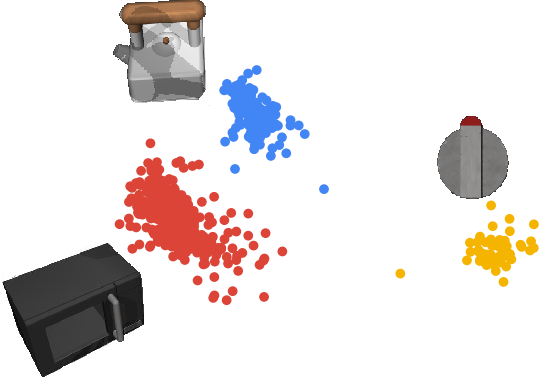}
        \caption{Diffusion prior}
        \label{fig:kitchenlatentvizdiffusion}
    \end{subfigure}
    \quad
    \begin{subfigure}[b]{0.3\linewidth}
        \centering
        \includegraphics[width=\linewidth]{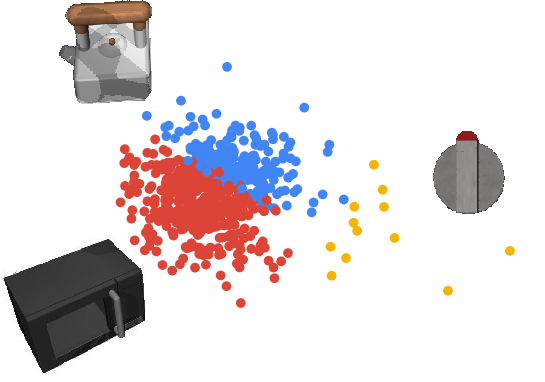}
        \caption{VAE prior}
        \label{fig:kitchenlatentvizvae}
    \end{subfigure}
    \caption{Visualization of latents projected using PCA for kitchen-mixed with $H=20$. The diffusion prior models the ground truth much more accurately while the VAE prior generates out-of-distribution samples.}
\end{figure}

In this section, we provide empirical evidence that latent diffusion models are superior in modelling multi-modal distributions as compared to VAEs.
For our experiment, we again consider the \textit{kitchen-mixed-v0} task. The goal of the generative model here is to learn the prior distribution $p(\bold{z} \mspace{-2mu}\mid\mspace{-2mu} \bold{s})$ and sample from it such that we can get candidate latents corresponding to state $\bold{s}$ belonging to the support of the dataset. However, as demonstrated earlier, the multi-modality in the latent spaces increases with the horizon. We visualize the latents from the initial states of all trajectories in the dataset in \textcolor{darkblue}{Figure} \ref{fig:kitchenlatentsvizgt} using PCA with $H = 20$.
The three clusters in the figure correspond to the latents of three different tasks namely microwave, kettle and burner.
Similarly, we also visualize the latents predicted by the diffusion model (\textcolor{darkblue}{Figure} \ref{fig:kitchenlatentvizdiffusion}) and the VAE conditional prior (\textcolor{darkblue}{Figure} \ref{fig:kitchenlatentvizvae}) for the same initial states by projecting them onto the principal components of the ground truth latents. We can see that the diffusion prior is able to sample effectively all modes from the ground truth latent distribution, while the VAE prior spreads its mass over the three modes, and thus samples out of distribution in between the three modes.
Using latents sampled from the VAE prior to learning the Q-function can thus lead to sampling from out of the support, leading to extrapolation error.

\subsection{Performance improvement with temporal abstraction}
\label{sec:performanceimprovement}
We empirically demonstrate the importance of temporal abstraction and the performance improvement with diffusion on modelling temporally abstract latent spaces. We compare our method with a variant of BCQ which uses temporal abstraction ($H>1$), which we refer to as BCQ-H. We use the same VAE architecture here as LDCQ, and fit the conditional Gaussian prior with a network having comparable parameters to our diffusion model.
\begin{wrapfigure}{r}{0.35\linewidth}\includegraphics[width=\linewidth]{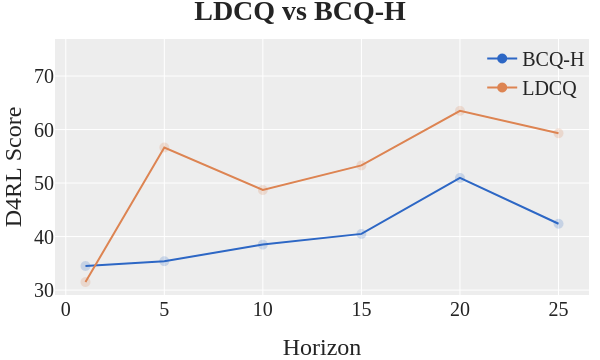}\caption{D4RL score of LDCQ and BCQ-H on kitchen-mixed-v0 with varying sequence horizon $H$}\label{diffusionvsbcq}\end{wrapfigure}
We find that generally, increasing the horizon $H$ results in better performance,
both in BCQ-H and LDCQ, and both of them eventually saturate and degrade, possibly due to the limited decoder capacity. With $H=1$, the latent distribution is roughly Normal as discussed earlier and our diffusion prior is essentially equivalent to the Gaussian prior in BCQ, so we see similar performance. As we increase $H$, however, the diffusion prior is able to efficiently sample from the more complex latent distribution that emerges, which allows the resulting policies to benefit from temporal abstraction. BCQ-H, while also seeing a performance boost with increased temporal abstraction, lags behind LDCQ. We plot D4RL score-vs-$H$ for BCQ-H and LDCQ evaluated on the \textit{kitchen-mixed-v0} task in \textcolor{darkblue}{Figure \ref{diffusionvsbcq}}.

\subsection{Offline RL benchmarks}
\label{sec:offlinerl}
In this section, we investigate the effectiveness of our Latent Diffusion Reinforcement Learning methods on the D4RL offline RL benchmark suite (\citet{fu2020d4rl}).
We compare with Behavior Cloning and several \textit{state-of-the-art} offline RL methods: Batch Constrained Q-Learing (BCQ) (\citet{fujimoto2019off}), Conservative Q-Learing (CQL) (\citet{kumar2020conservative}), Implicit Q-Learning (IQL) (\citet{kostrikov2021offline}), Decision Transformer (DT) (\citet{Chen2021DecisionTR}), Diffuser (\citet{janner2022planning}) and Decision Diffuser (\citet{ajay2022conditional}). The last two algorithms are previous trajectory diffusion methods. We found that our method does not require much hyperparameter tuning and only had to vary the sequence horizon $H$ across tasks. In maze2d and AntMaze tasks we use $H=30$, in kitchen tasks we use $H=20$ and in locomotion tasks we use $H=10$. We train our diffusion prior with $T=200$ diffusion steps. The other hyperparameters which are constant across tasks are provided in the supplemental material.
In Table \ref{tab:mazekitchen}, we show results on the sparse-reward tasks in D4RL which require long horizon trajectory stitching. In particular, we look at tasks in Maze2d, AntMaze and FrankaKitchen environments which are known to be the most difficult in D4RL, with most algorithms performing poorly. Maze2d and AntMaze consist of undirected demonstrations controlling the agent to navigate to random locations in a maze.
AntMaze is quite difficult because the agent must learn the high-level trajectory stitching task alongside low-level control of the ant robot with 8-DoF. %
In the maze navigation tasks, we also evaluate the performance of our goal-conditioned (LDGC) variant. For Diffuser runs we use the goal-conditioned inpainting version proposed by the authors since the classifier-guided version yielded poor results. We found our implementation of BCQ improved over previous reported scores in kitchen tasks.
\begin{table}[htpb]
    \caption{Performance comparison on D4RL tasks which require long-horizon stitching with high multimodality. Goal conditioning (LDGC) variant is evaluated in the navigation environments.}
  \centering
    \resizebox{\columnwidth}{!}{\begin{tabular}{lccccccccc}
    \toprule
    \textbf{Dataset} & \textbf{BC}   &
    \textbf{BCQ}   &
    \textbf{CQL}   & \textbf{IQL} & \textbf{DT} & \textbf{Diffuser} & \textbf{DD} & \textbf{LDCQ (Ours)} & \textbf{LDGC (Ours)}
    \\
    \midrule
    maze2d-large-v1 & 5.0& 6.2 & 12.5& 58.6& 18.1&123.0& -& \textbf{150.1} $\pm$ 2.9 & \textbf{206.8} $\pm$ 3.1
    \\
    \midrule
    antmaze-medium-diverse-v2 & 0.0& 0.0 & 53.7& \textbf{70.0}& 0.0& 45.5& 24.6& \textbf{68.9} $\pm$ 0.7 & \textbf{75.6} $\pm$ 0.9
    \\
    antmaze-large-diverse-v2 & 0.0& 2.2 & 14.9& 47.5& 0.0& 22.0& 7.5& \textbf{57.7} $\pm$ 1.8 & \textbf{73.6} $\pm$ 1.3
    \\
    \midrule
    kitchen-partial-v0 & 38.0& 31.7 &50.1& 46.3& 42.0& -& 57.0& \textbf{67.8} $\pm$ 0.8 & -\\
    kitchen-mixed-v0 & 51.5& 34.5 & 52.4& 51.0& 50.7& -& \textbf{65.0} & \textbf{62.3} $\pm$ 0.5 & -\\
    \bottomrule
    \end{tabular}}

  \label{tab:mazekitchen}%
\end{table}
Both our methods (LDCQ and LDGC) achieve state-of-the-art results in all sparse reward D4RL tasks. The goal-conditioned variant outperforms all others in maze2d and AntMaze. This variant is extremely simple to implement through supervised learning of the diffusion prior with no Q-learning or online planning and is ideal for goal-reaching tasks. %
We also provide an evaluation of our method on the D4RL locomotion suite (Table \ref{tab:locomotion}) and the Adroit robotics suite . While these tasks are not specifically focused on trajectory-stitching, our method is competitive with other offline RL methods. We only run the LDCQ variant here since they are not goal-reaching tasks.

\begin{table}[htpb]
      \caption{Performance comparison on the D4RL locomotion tasks.}
  \centering
    \resizebox{\columnwidth}{!}{\begin{tabular}{lcccccccc}
    \toprule
    \textbf{Dataset} & \textbf{BC}   & 
    \textbf{BCQ}   &
    \textbf{CQL}   & \textbf{IQL} & \textbf{DT} & \textbf{Diffuser} & \textbf{DD} & \textbf{LDCQ (Ours)} \\
    \midrule
    halfcheetah-medium-expert-v2 & 55.2& 64.7 & \textbf{91.6}& 86.7& 86.8& 88.9& \textbf{90.6}& \textbf{90.2} $\pm$ 0.9\\
    walker2d-medium-expert-v2 & 107.5& 57.5 & \textbf{108.8}& \textbf{109.6} & 108.1& 106.9& \textbf{108.8}& \textbf{109.3} $\pm$ 0.4\\
    hopper-medium-expert-v2 & 52.5& \textbf{110.9} &105.4& 91.5& 107.6& 103.3& \textbf{111.8}& \textbf{111.3} $\pm$ 0.2\\
    \midrule
    halfcheetah-medium-v2 & 42.6& 40.7 & 44.0& 47.4& 42.6& 42.8& \textbf{49.1}&42.8 $\pm$ 0.7\\
    walker2d-medium-v2 & 75.3& 53.1 & 72.5& 78.3& 74.0& 79.6& \textbf{82.5}& 69.4 $\pm$ 3.5\\
    hopper-medium-v2 & 52.9& 54.5 & 58.5& 66.3& 67.6& 74.3& \textbf{79.3}& 66.2 $\pm$ 1.7\\
    \midrule
    halfcheetah-medium-replay-v2 &36.6 & 38.2 &\textbf{45.5} &\textbf{44.2} &36.6 &37.7 & 39.3& 41.8 $\pm$ 0.4\\
    walker2d-medium-replay-v2 & 26.0 & 15.0 & \textbf{77.2}& 73.9& 66.6& 70.6& 75.0& 68.5 $\pm$ 4.3\\
    hopper-medium-replay-v2 & 18.1 & 33.1 & 95.0& 94.7& 82.7& 93.6& \textbf{100.0}& 86.2 $\pm$ 2.5\\
    \bottomrule
    \end{tabular}}

  \label{tab:locomotion}%
\end{table}
\begin{table}[!htpb]
    \caption{Performance comparison on Adroit tasks.}
  \centering
    {\begin{tabular}{l|ccccccc|ccc}
    \toprule
    \textbf{Dataset} & \textbf{BC}   &
    \textbf{BCQ}   &
    \textbf{CQL}   & \textbf{IQL} &
    \textbf{DT} & 
    \textbf{Diffuser} & 
    \textbf{DD} &
    \textbf{LDCQ (Ours)}
    \\
    \midrule
    pen-human & 34.4 & 68.9 & 37.5 & 71.5 &
    -& 
    - &
    -  &
    \textbf{74.1}  \\
    hammer-human & 1.2 & 0.3 & \textbf{4.4} & 1.4 &
    -& 
    - &
    -  &
    1.5  \\
    door-human & 0.5 & 0.0 & 9.9 & 4.3 &
    -& 
    - &
    -  &
    \textbf{11.8}  \\
    relocate-human & 0.0 & -0.1 & 0.2 & 0.1 &
    -& 
    - &
    -  &
    \textbf{0.3}  \\
    pen-cloned & 37.0 & 44.0 & 39.2 & 37.3 & - & -& - & \textbf{47.7} \\
    hammer-cloned & 0.6 & 0.4 & 2.1 & 2.1 & - & - & - & \textbf{2.8}\\
    door-cloned & 0.0 & 0.0 & 0.4 & \textbf{1.6} & -&- & - & 1.1\\
    relocate-cloned & -0.3 & -0.3 & -0.1 & -0.2 & - & - & - & -0.2\\
    \bottomrule
    \end{tabular}}

  \label{tab:adroit}%
\end{table}

To extend our method for tasks with high-dimensional image input spaces, we propose to compress the image space using a $\beta$-VAE encoder such that our method operates on a compressed state space. We present the results on the CARLA D4RL task in table \ref{tab:carla}. (Further details in Appendix \ref{carla}).

\begin{table}[htpb]
    \caption{Performance comparison on image-based CARLA task.}
  \centering
    \resizebox{0.7\columnwidth}{!}{\begin{tabular}{l|ccccccc|ccc}
    \toprule
    \textbf{Dataset} & \textbf{BC}   &
    \textbf{BCQ}   &
    \textbf{CQL}   & \textbf{IQL} & \textbf{DT} & \textbf{Diffuser} & \textbf{DD} & \textbf{LDCQ}
    \\
    \midrule
    carla-lane-v0 & 17.2 & -0.1 & 20.9 & 18.6 & -& - & -  & \textbf{24.7}  \\
    \bottomrule
    \end{tabular}}
  \label{tab:carla}%
\end{table}

\vspace{-10pt}
\section{Limitations}
Our method, like other diffusion-based RL algorithms is slow at inference time due to the iterative sampling process, especially since we use a simple implementation of DDPM. This could be mitigated with methods that can perform faster sampling (\citet{song2020denoising}, \citet{lu2022dpm}, \citet{dockhorn2022score}, \citet{xiao2022DDGAN}), or by distilling these diffusion models into others methods which need fewer sampling steps (\citet{song2023consistency}, \citet{salimansprogressive}).
Our method has average performance on the locomotion task suite while having significant gains in the sparse reward tasks. We suspect the high periodicity of the walking gaits in the locomotion suite does not benefit much from reasoning with temporal abstraction. We also do not use a perturbation function during Q-learning like \citet{fujimoto2019off}, which makes it difficult for us to improve over the poor controllers in medium and medium-replay locomotion datasets. Introducing a perturbation function requires careful tuning to avoid extrapolation error, and the converged Q-learning wouldn't necessarily correspond to a high value policy, which is why other offline RL methods, which try to balance this tradeoff, evaluate online during training and consider the best scores. We however only evaluate a policy once after training is fully complete.
Another shortcoming of our work is that the sequence horizon $H$ for temporal abstraction has to be fixed for the entire experiment. We expect that being able to vary this adaptively could improve performance.
\vspace{-8pt}

\section{Conclusion}
In this work, we showed that offline RL datasets comprised of suboptimal demonstrations have expressive multi-modal latent spaces which can be captured with temporal abstraction and is well suited for learning high-reward policies. With a powerful conditional generative model to capture the richness of this latent space, we demonstrated that the simple batch-constrained Q-learning framework can be directly used to obtain strong performance. Our biggest improvements come from long-horizon sparse reward tasks, which most prior offline RL methods struggled with, even previous raw trajectory diffusion methods. Our approach also required no task-specific tuning, except for the sequence horizon $H$. We believe that latent diffusion has enormous potential in offline RL and our work has barely scratched the surface of possibilities.

\bibliographystyle{plainnat}
\bibliography{refs}
\newpage
\renewcommand*{\thesection}{\Alph{section}}
\setcounter{section}{0}

\section{Training Details}

\subsection{Source Code}
The source code is available at: \href{https://github.com/ldcq/ldcq}{\texttt{https://github.com/ldcq/ldcq}}.

\subsection{Network Architecture}
\label{sec:app:network-architectures}
\subsubsection{Variational Autoencoder}
\textbf{Encoder.} For learning the latent trajectory representation, our VAE uses an architecture similar to \citet{ajay2020opal}.
The encoder consists of two stacked bidirectional GRU layers, followed by mean and standard deviation heads which are each a 2 layer MLP with RELU activation for the hidden layers. The mean  output head is a linear layer. The standard deviation output head is followed by a SoftPlus activation function to ensure it is always positive. The hidden layer dimension is fixed to 256.

\textbf{Decoder.} For the low-level policy decoder, we use an auto-regressive policy network similar to that described in EMAQ (\citet{ghasemipour2021emaq}), in which each element of the action vector has its own MLP network, taking as input the current state, latent representation, and all previously-sampled action elements. The complete action vector is sampled element-by-element, with the most recently sampled element becoming an input to the network for the next element. These MLP networks consists of 2 layers followed by 2 layers of mean and standard deviation heads similar to the encoder network. The mean output head is a linear layer and the standard deviation output head is followed by a SoftPlus activation.  Again, ReLU activation is used after all hidden layer and the hidden dimension is fixed to 256.
\subsubsection{Diffusion Prior}
The diffusion prior is a deep ResNet (\citet{he2016deep})
architecture consisting of 8 residual blocks. It takes as input a vector representing a latent trajectory $\bold{z}$ and outputs a denoised version of the latent. The hidden blocks are of dimensions: [128, 32, 16, 8, 16, 32, 128]. Similar to a U-Net (\citet{inproceedings}),
the initial blocks are connected by residual connections to the later blocks having the same hidden dimension. The diffusion timestep $t$ is encoded with a 256-dimensional sinusoidal embedding and then further encoded with a 2-layer MLP. The conditioning state $\bold{s}$ is also encoded by a 2 layer MLP. In each residual block, the state and time encodings are concatenated with the current layer activation for conditioning. When training the unconditional diffusion model for classifier-free guidance, the state input is given as a vector of zeros to represent a null vector.

\subsubsection{Q-networks}
The Q-networks take as input the state $\bold{s}$, latent $\bold{z}$ and consist of a 5 layer MLP with 256 hidden units in the first 3 layers, 32 hidden units in the third layer, and finally a linear output layer. We use GELU activation function between hidden layers. LayerNorm is applied before each activation.

\subsection{Hyperparameters}
The hyperparameters which are constant across tasks for the different stages of our proposed method are given in Tables \ref{tab:vae}, \ref{tab:diffusion} and \ref{tab:qnet}.

\begin{table}[!htpb]
    \caption{$\beta$-VAE hyperparameters}
  \centering
    {\begin{tabular}{lcc}
    \toprule
    \textbf{Parameter} & \textbf{Value}
    \\
    \midrule
    Learning rate & 5e-5\\
    Batch size & 128\\
    Epochs & 100\\
    Latent dimension ($\bold{z}$) & 16\\
    $\beta$ & 0.05\\
    Hidden layer dimension & 256\\
    \bottomrule
    \end{tabular}}
    \label{tab:vae}
\end{table}

\begin{table}[!htpb]
    \caption{Diffusion training hyperparameters}
  \centering
    {\begin{tabular}{lcc}
    \toprule
    \textbf{Parameter} & \textbf{Value}
    \\
    \midrule
    Learning rate & 1e-4\\
    Batch size & 32\\
    Epochs & 300\\
    Diffusion steps ($T$) & 500\\
    Drop probability (For unconditional prior) & 0.1\\
    Variance schedule & linear\\
    Sampling algorithm & DDPM\\
    $\gamma$ (For Min-SNR-$\gamma$ weighing) & 5\\
    \bottomrule
    \end{tabular}}
    \label{tab:diffusion}
\end{table}

\begin{table}[!htpb]
    \caption{Q-Learning hyperparameters}
  \centering
    \resizebox{\columnwidth}{!}{\begin{tabular}{lcc}
    \toprule
    \textbf{Parameter} & \textbf{Value}
    \\
    \midrule
    Learning rate & 5e-4\\
    Batch size & 128\\
    Discount factor ($\gamma$) & 0.995\\
    Target net update rate ($\rho$) & 0.995\\
    PER buffer $\alpha$ & 0.7\\
    PER buffer $\beta$ & Linearly increased from 0.3 to 1, Grows by 0.03 every 3000 steps\\
    Diffusion samples for batch argmax & 500\\
    \bottomrule
    \end{tabular}}
    \label{tab:qnet}
\end{table}

\subsection{Hardware}
The models were trained on NVIDIA RTX A6000. Since different tasks have different dataset sizes, the model training times changes across tasks. Depending on the task, training the $\beta$-VAE took between 3-7 hours, the diffusion prior took between 4-12 hours and the Q-Learning took between 3-5 hours.

\section{Latent Diffusion-Constrained Planning (LDCP)}
In this section, we explore another method to derive a policy for offline RL with latent diffusion other than our proposed methods \textit{Latent Diffusion-Constrained Q-Learning (LDCQ)} and \textit{Latent Diffusion Goal Conditioning (LDGC)}.
This is a model-based method which learns a temporally abstract world model of the environment from offline data. Specifically, we learn a temporally abstract world model $p_{\eta}(\bold{s}_{t + H} \mid \bold{s}_t, \bold{z})$ that predicts the state outcome of executing a particular latent behavior after $H$ steps. That is, given the current state $\bold{s}_t$ and a latent behavior $\bold{z}$ the model predicts the distribution of the state $\bold{s}_{t + H}$. This is trained in a supervised manner by sampling transition tuples $(\bold{s}_t, \bold{z}, \bold{s}_{t + H})$ from $\boldsymbol{\tau}_H \sim \mathcal{D}$ and minimizing the objective:
\begin{equation}
    \mathcal{L}(\eta) = \mathbb{E}_{\boldsymbol{\tau}_H \sim \mathcal{D}} \mid \mid p_{\eta}(\bold{s}_{t + H} \mid \bold{s}_t, \bold{z}) - \bold{s}_{t + H} \mid \mid ^2
\end{equation}
where $\eta$ are the parameters of the temporally abstract world model $p_\eta$.

In goal-reaching environments, we leverage this model to do planning using the diffusion prior. We sample $n$ latents $\bold{z}^i$ (1 $\leq$ $i$ $\leq$ $n$) using the diffusion prior for the current state $\bold{s}_t$, and use the learnt dynamics model to compute predicted future state $\bold{s}_{t+H}^i$ for each latent $\bold{z}^i$. These final states are then scored using a cost-function $\mathcal{J}$ and the latent corresponding to the best final state is chosen for execution. 
Note that sampling latents from the diffusion prior ensures that the world model is not queried on out-of-support data. We refer to this method as \textit{Latent Diffusion-Constrained Planning (LDCP)}.
The planning procedure is summarized in Algorithm \ref{alg:ldcp}.

\begin{algorithm}[htpb]
\caption{Latent Diffusion-Constrained Planning (LDCP)}
\label{alg:ldcp}
\begin{algorithmic}[1]
\State \textbf{Input:} horizon $H$, number of latents to sample $n$, maximum iterations $M$, cost-function $\mathcal{J}$, policy decoder $\pi_\theta$, temporally abstract world model $p_\eta$, latent diffusion denoising function $\mu_\psi$, variance schedule $\alpha_1, \dots , \alpha_T$, $\bar{\alpha}_1, \dots , \bar{\alpha}_T$, $\beta_1, \dots , \beta_T$.

\State \textit{done} = \textit{False}
\While{not \textit{done}}
    \State Observe environment state $\bold{s}_0$

    \State Sample $n$ latents: $\bold{z}_T \sim \mathcal{N(\bold{0}, \bold{I})}$

    \For{$t=T$ to $1$} \Comment{DDPM Sampling}
    \State $\hat{\bold{z}} = \mu_\psi(\bold{z}_t, $\O{}$, t) + w (\mu_\psi(\bold{z}_t, \bold{s}_0, t) - \mu_\psi(\bold{z}_t, $\O{}$, t))$
    
    \State $\bold{z}_{t-1} \sim \mathcal{N}(\frac{\sqrt{\alpha_t}(1-\bar{\alpha}_{t-1})}{1 - \bar{\alpha}_t} \bold{z}_t + \frac{\sqrt{\bar{\alpha}_{t-1}}\beta_t}{1 - \bar{\alpha}_t} \hat{\bold{z}},
    \mathbb{I} (t > 1)\beta_t\bold{I})$
    
    \EndFor

    \State Compute future states for each latent $\bold{z}_0^i$: $\bold{s}_{H}^i = p_\eta(\bold{s}_H^i \mid \bold{s}_0, \bold{z}_0^i)$

    \State Find best latent based on the cost-function: $i = \underset{i}{\operatorname{argmin}} 
 \hspace{0.5em} \mathcal{J}(\bold{s}_H^i)$

 \State Compute action-sequence using policy decoder $\pi_\theta (\bold{a} \mid \bold{s}_0, \bold{z}_0^i)$

 \State $h = 0$
 \While {$h < H \textbf{and}$ not \textit{done}}
    \State Execute action $\bold{a}_h$
    \State Update \textit{done}
    \State $h = h + 1$
 \EndWhile

\EndWhile
\end{algorithmic}
\end{algorithm}

The cost-function which we use for the goal-reaching environments is the Euclidean distance to the goal. We can also extend this planning to horizons greater than $H$ by further sampling latents for each future state $\bold{s}_{t + H}^i$ (1 $\leq$ $i$ $\leq$ $n$). This means, after sampling $n$ latents for $\bold{s}_t$ with the diffusion prior, we further sample $n$ 
more latents for each of the future states $\bold{s}_{t + H}^i$. This increases the `planning depth' $d$. The final states at the last level of planning are then scored using the cost-function and the latent at the first level which led to that final state is chosen for execution. This procedure complexity grows exponentially and thus the planning depth has to be restricted. For a planning depth of $d$, there are $n^d$ model calls. We found a planning-depth of $d=2$ to be sufficient for all navigation environments achieving state-of-the-art results. Thus, with a latent sequence horizon of $H=30$, our total planning horizon is 60. The results are tabulated in Table \ref{tab:ldcp}.

\begin{table}[htpb]
    \caption{Performance comparison on D4RL navigation tasks with LDCP.}
  \centering
    \resizebox{\columnwidth}{!}{\begin{tabular}{lcccccccccc}
    \toprule
    \textbf{Dataset} & \textbf{BC}   &
    \textbf{BCQ}   &
    \textbf{CQL}   & \textbf{IQL} & \textbf{DT} & \textbf{Diffuser} & \textbf{DD} &
    \textbf{LDCQ (Ours)} &
    \textbf{LDGC (Ours)} &
    \textbf{LDCP (Ours)}
    \\
    \midrule
    maze2d-large-v1 & 5.0& 6.2 & 12.5& 58.6& 18.1&123.0& -& 
    \textbf{150.1} $\pm$ 2.9 &
    \textbf{206.8} $\pm$ 3.1 &
    \textbf{184.3} $\pm$ 3.8
    \\
    \midrule
    antmaze-medium-diverse-v2 & 0.0& 0.0 & 53.7& 70.0 & 0.0& 45.5& 24.6&
    \textbf{68.9} $\pm$ 0.7 & 
    \textbf{75.6} $\pm$ 0.9 &
    \textbf{77.0} $\pm$ 1.1
    \\
    antmaze-large-diverse-v2 & 0.0& 2.2 & 14.9& 47.5& 0.0& 22.0& 7.5& 
    \textbf{57.7} $\pm$ 1.8 & 
    \textbf{73.6} $\pm$ 1.3 &
    \textbf{59.7} $\pm$ 1.3
    \\
    \bottomrule
    \end{tabular}}
  \label{tab:ldcp}%
\end{table}

\subsection{Visualizing Model Predictions}
Learning a world model also allows us to visualize the effect of executing any given latent behavior. This means, even when the model is not used for planning, like in LDCQ, it can be used to compute the final state that will be reached for every latent behavior from a particular state. This information can be used to understand if the model is learning reasonable behavior modes.

 \begin{figure}[htpb]

        \centering
        \includegraphics[width=0.5\linewidth]{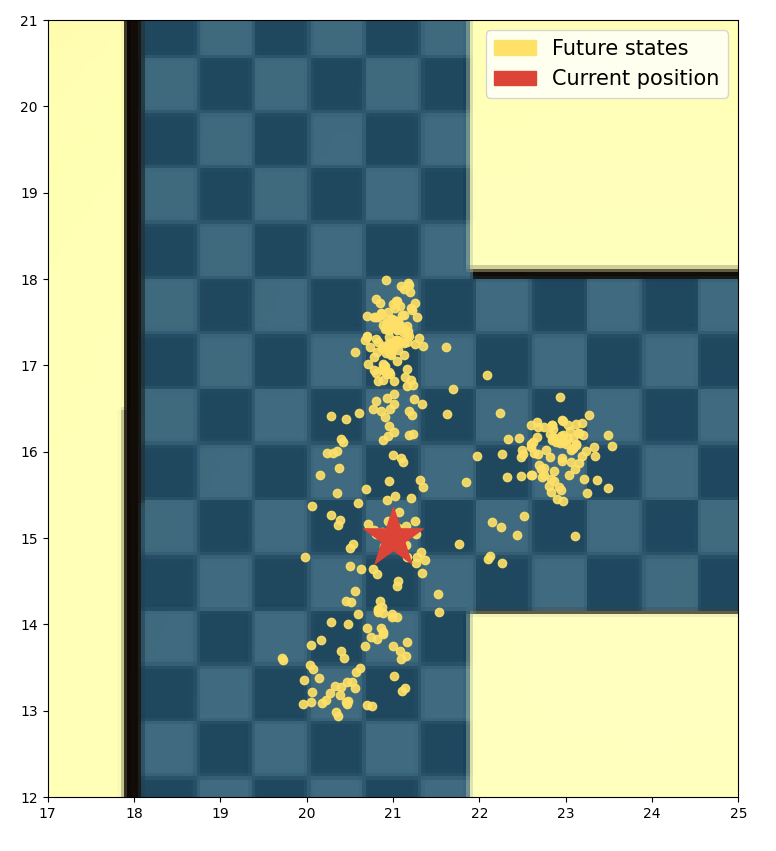}

\caption{\textbf{Visualizing model predictions:} Visualization of future states with latents sampled from the diffusion prior at a T-intersection in antmaze-large-diverse-v2 D4RL task. We can see multimodal future state predidctions corresponding to 3 possible directions at the T-intersection.}
    \label{fig:interpretability}
\end{figure}

We plot the \textit{xy}-coordinates of our abstract world model $p_\eta(\bold{s}_{t + H} \mspace{-2mu}\mid\mspace{-2mu} \bold{s}_{t}, \bold{z})$ predictions at a \textit{T}-intersection of AntMaze large environment for latents sampled from our diffusion prior $\bold{z} \sim p_\psi(\bold{z} \mspace{-2mu}\mid\mspace{-2mu} \bold{s}_t)$ in \textcolor{darkblue}{Figure \ref{fig:interpretability}} to demonstrate this. The plot shows that the diffusion prior sampled latents which go in all the three directions at the T-intersection.

\section{CARLA Autonomous Driving task}
\label{carla}
\begin{figure}[htpb]
    \centering
\includegraphics[width=\linewidth]{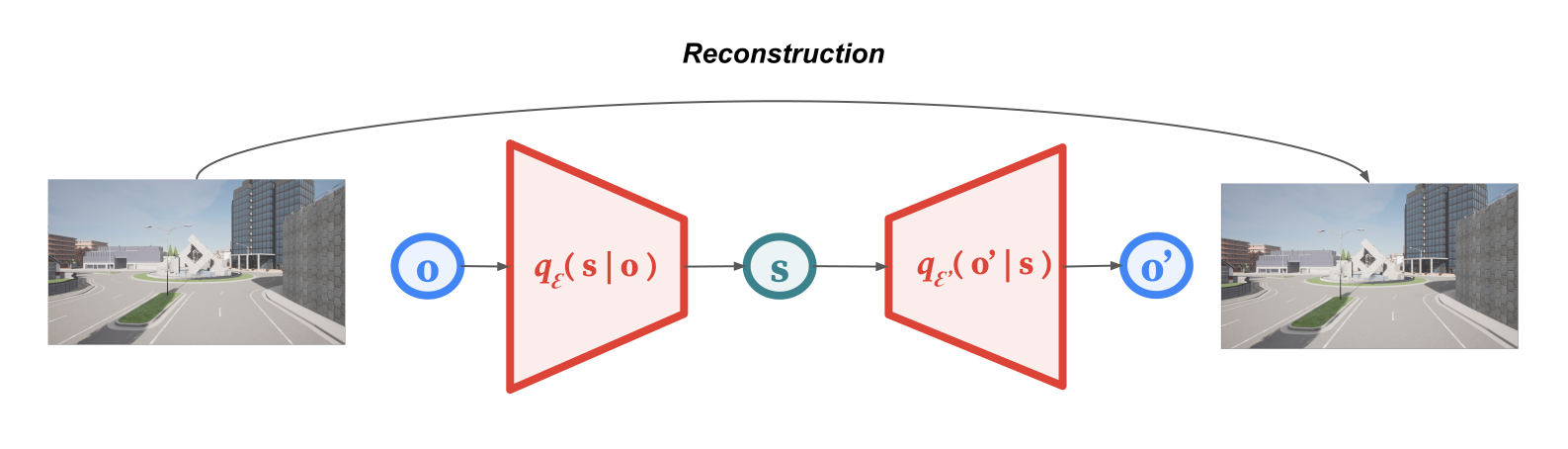}
    \caption{Autoencoder training for image-based task}
    \label{fig:carla}
\end{figure}

To extend our method for tasks with high-dimensional image input spaces, we propose to compress the image space such that our method operates on this compressed state space. We create a low-dimensional compressed representation using an autoencoder $\mathcal{E}$ before using the LDCQ framework. Note that this encoder operates on a single image and not on a temporal sequence of images (\textcolor{darkblue}{Figure} \ref{fig:carla}). The downstream framework of LDCQ however operates on the temporal compressed image sequences.

We evaluate the performance of our method on the CARLA autonomous driving D4RL task. The task consists of an agent which has control to the throttle (gas pedal), the steering, and the break pedal for the car. It receives 48 $\times$ 48 RGB images from the driver’s perspective as observations. We use a $\beta$-VAE architecture to create a 32-dimensional compressed state for this task. The horizon for LDCQ is set to $H=30$.

\section{Random walk 1D}

 In this experiment, we construct a simple toy problem to show how sampling effectively from the multimodal behavioral distribution helps the diffusion prior outperform a Gaussian VAE prior during Q-learning. We construct a simple toy MDP with a one-dimensional state space $\mathcal{S}\in[-10.0,10.0]$. The agent starts at the origin (0,0) and receives a reward of 10 if it reaches either the far left (-10.0) or far right (10.0) state, and -1 reward every other step. The environment times out after 500 steps. The action is the distance moved in that timestep with a max distance of length 1, $\mathcal{A} \in[-1.0, 1.0]$. The dataset consists of episodes where the agent randomly selects actions from the uniform distribution $a \sim \mathcal{U}([-1.0,-0.8]\cup[0.8,1])$. This means the agent has a step size between 0.8 to 1.0 units either left or right every timestep. We train a VAE to try to fit this action distribution, and use BCQ to learn a policy. We also train train a diffusion based policy with LDCQ, using $H=1$ and compare the results.

 \begin{figure}[htpb]
    \centering
\includegraphics[width=0.7\textwidth]{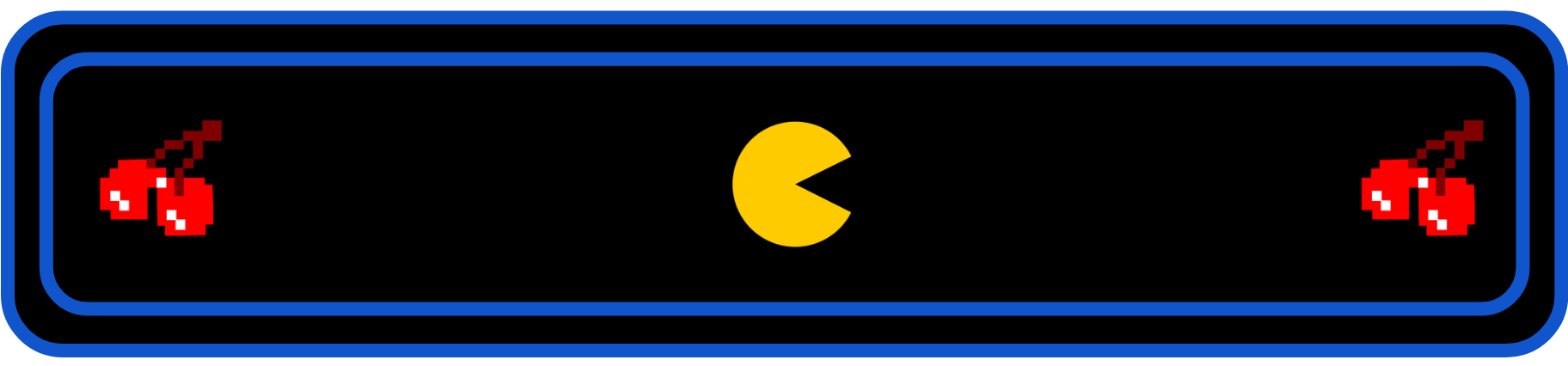}
    \caption{1D Random walk}
\end{figure}

We find that the VAE frequently samples actions never present in the dataset. This is because the Gaussian mean to the above action distribution is 0.0, but no actual actions are present between $(-0.8,0.8)$ where a large proportion of probability mass is assigned by the Gaussian model. 
Meanwhile, the diffusion prior is able to fit the 2 modes quite well. After 10000 steps of Q-learning, the diffusion constrained policy learns to navigate to either end perfectly and achieves an average reward of $\textbf{-2.2}$ while the VAE constrained policy is still almost random, frequently taking actions with small step size and an average reward of $\textbf{-66}$.

\section{Increasing diffusion steps improves performance}
\begin{figure}[ht]
    \centering
    \begin{subfigure}[b]{0.45\linewidth}
        \centering
        \includegraphics[width=\linewidth]{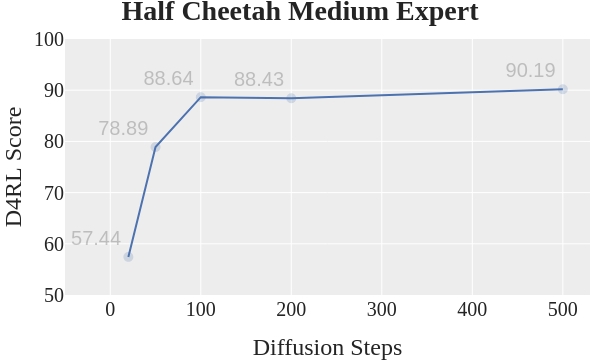}
        \label{fig:halfcheetahdiffusion}
    \end{subfigure}
    \quad
    \begin{subfigure}[b]{0.45\linewidth}
        \centering
        \includegraphics[width=\linewidth]{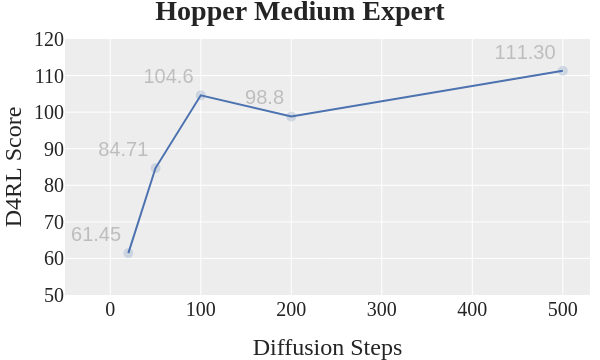}
        \label{fig:hopperdiffusion}
    \end{subfigure}
    \quad
    \begin{subfigure}[b]{0.45\linewidth}
        \centering
        \includegraphics[width=\linewidth]{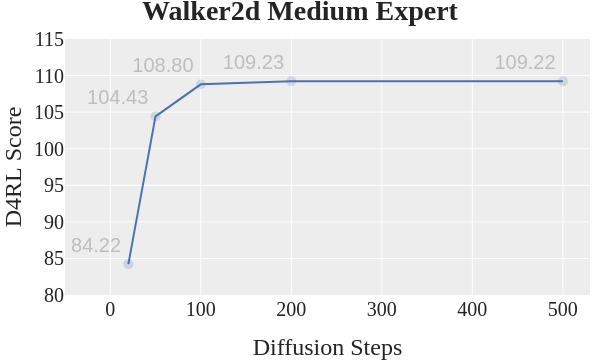}
        \label{fig:walker2ddiffusion}
    \end{subfigure}
    \caption{D4RL score for LDCQ with varying diffusion steps $T$ in locomotion tasks. }
    \label{fig:timesteps}
\end{figure}

We study the impact of the number of diffusion steps on the performance for LDCQ.
We found that for the locomotion tasks, increase in diffusion timesteps $T$ during evaluation generally corresponds to increase in task performance. We plot these results in Figure \ref{fig:timesteps}.

\begin{figure}[ht]
    \centering
        \includegraphics[width=0.6\linewidth]{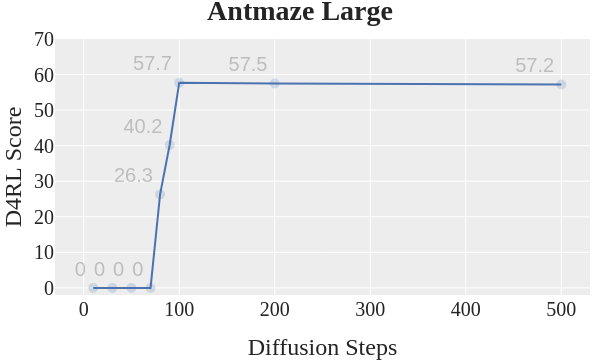}
    \caption{D4RL score for LDCQ with varying diffusion steps $T$}
    \label{fig:antmazediffusion}
\end{figure}

For the long horizon tasks, we found that increasing diffusion steps resulted in an initial trend upward in performance. Beyond this, the performance does not improve with additional diffusion steps (\textcolor{darkblue}{Figure} \ref{fig:antmazediffusion}).

We also used additional diffusion steps at time $t=0$ similar to Diffusion-X (\citet{pearce2023imitating}). This means that after the DDPM sampling of diffusion from time $T$ to 1, we run $X$ additional diffusion steps to further denoise the sample, assuming time-step $t=1$. \citet{pearce2023imitating} reasoned that this pushes the samples further towards higher-likelihood regions. We used 10 additional steps across experiments and found this to slightly improves performance.

\section{Performance improvement with temporal abstraction}

\begin{figure}[!htpb]
    \centering
    \begin{subfigure}[b]{0.45\linewidth}
        \centering
        \includegraphics[width=\linewidth]{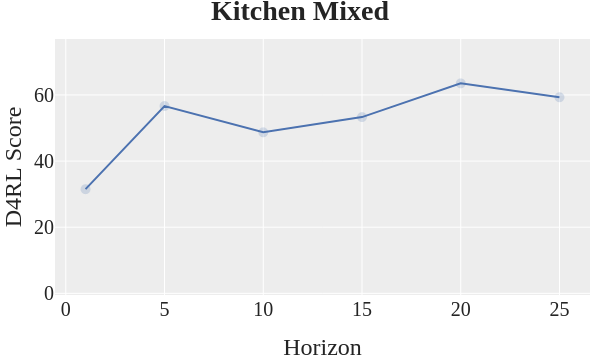}
        \label{fig:kitchenmixedhorizon}
    \end{subfigure}
    \quad
    \begin{subfigure}[b]{0.45\linewidth}
        \centering
        \includegraphics[width=\linewidth]{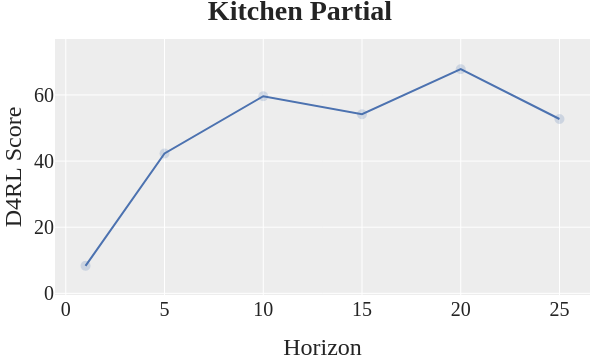}
        \label{fig:kitchenpartialhorizon}
    \end{subfigure}
    \quad
    \begin{subfigure}[b]{0.45\linewidth}
        \centering
        \includegraphics[width=\linewidth]{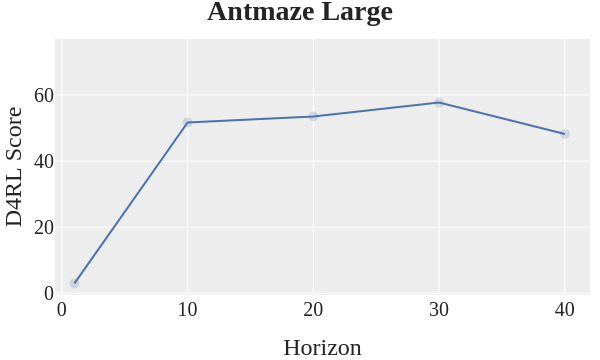}
        \label{fig:antmazelargehorizon}
    \end{subfigure}
    \quad
    \begin{subfigure}[b]{0.45\linewidth}
        \centering
        \includegraphics[width=\linewidth]{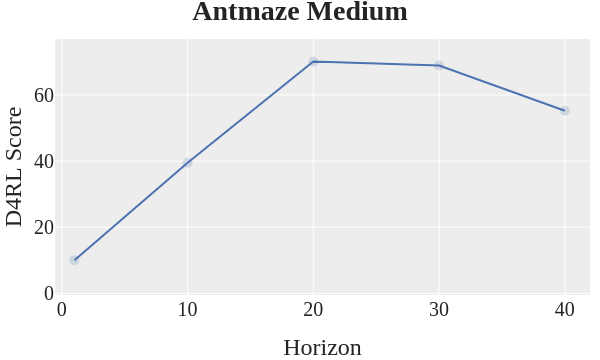}
        \label{fig:antmazemediumhorizon}
    \end{subfigure}
    \quad
    \caption{D4RL score for LDCQ with varying sequence horizons $H$. }
    \label{fig:horizons}

\end{figure}

We provided empirical evidence for improvement in performance as we increase temporal abstraction or horizon $H$ for the \textit{kitchen-mixed-v0} environment. We see similar trends for the other long-horizon tasks as well (\textcolor{darkblue}{Figure} \ref{fig:horizons}). The performance in general improves with increasing temporal abstraction but beyond a certain point, it drops possibly because of the limited capacity of the policy decoder.

For the locomotion tasks, we did not observe any noticeable difference with increase in temporal abstraction, so we ended up using a moderate sequence length $H=10$. This could be due to the high frequency periodicity of these tasks that does not require much look-ahead.

\end{document}